%% file: causalBandit_main.tex
\documentclass{article}

\usepackage[english]{babel}
\usepackage{amsfonts}
\usepackage{amsmath}
\usepackage{times}

\usepackage{graphicx} 
\usepackage{subfigure} 
\usepackage{booktabs} 

\usepackage{algorithm}
\usepackage{algorithmic}
\usepackage{amsthm}

\usepackage{hyperref}
\usepackage{url}
\usepackage{natbib}

\DeclareMathOperator*{\argmax}{arg\,max}

\newcommand{\bigCI}{\mathrel{\text{\scalebox{1.07}{$\perp\mkern-10mu\perp$}}}}

\usepackage[accepted]{icml2018}

\usepackage{comment}

\begin{document} 

\twocolumn[
\icmltitle{Contextual Multi-Armed Bandits for Causal Marketing}
\icmltitlerunning{Contextual Multi-Armed Bandits for Causal Marketing}



\icmlsetsymbol{equal}{*}

\begin{icmlauthorlist}
\icmlauthor{Neela Sawant}{Amazon}
\icmlauthor{Chitti Babu Namballa}{Amazon}
\icmlauthor{Narayanan Sadagopan}{Amazon}
\icmlauthor{Houssam Nassif}{Amazon}
\end{icmlauthorlist}

\icmlaffiliation{Amazon}{Amazon.com, USA}

\icmlcorrespondingauthor{Neela Sawant}{nsawant@amazon.com}

\icmlkeywords{Causal inference, Conditional Average Treatment Effect, Multi-Arm Bandits, Thompson Sampling}

\vskip 0.3in
]



\printAffiliationsAndNotice{}  

\begin{abstract}
  \input{Abstract}
\end{abstract}

\input{Introduction}

\input{RelatedWork}

\input{CausalProposal}

\input{Experiments}

\input{Results}

\input{Conclusion}


\bibliography{../bibliography/hvaeReferences}
\bibliographystyle{icml2018}

\appendix
\input{Appendix}

\end{document}

%% file: Abstract.tex
This work explores the idea of a \emph{causal contextual multi-armed bandit} approach to automated marketing, where we estimate and optimize the causal (incremental) effects. Focusing on causal effect leads to better return on investment (ROI) by targeting only the persuadable customers who wouldn't have taken the action organically. Our approach draws on strengths of causal inference, uplift modeling, and multi-armed bandits. It optimizes on causal treatment effects rather than pure outcome, and incorporates counterfactual generation within data collection. Following uplift modeling results, we optimize over the incremental business metric. Multi-armed bandit methods allow us to scale to multiple treatments and to perform off-policy policy evaluation on logged data. The Thompson sampling strategy in particular enables exploration of treatments on similar customer contexts and materialization of counterfactual outcomes. Preliminary offline experiments on a retail Fashion marketing dataset show merits of our proposal.

%% file: Introduction.tex
\section{Introduction}
Personalizing marketing campaigns is a focal problem in advertising~\cite{Chen2009}. Conventional advertising and recommendation approaches optimize on observed outcomes such as clicks and transactional revenue that tend to reinforce known behaviors~\cite{Teo2016airstream, Nassif2016DiversifyingMR}. For instance, if a customer is already a heavy purchaser in mobile apps, a model optimizing on observed conversions may show more mobile app marketing campaigns to the customer, even if the customer would have made the purchase anyway. In fact, a recent study found that only about a quarter of the clicks typically attributed to the recommender system on Amazon are actually caused by it~\cite{Sharma:2015:ECI:2764468.2764488}. 

To improve the return on investment (ROI), there is a need to optimize for incremental (or causal) effects~\cite{rubin_waterman} in personalizing marketing campaigns for customers\footnote{We use following words interchangeably - (a) campaign, treatment; (b) incremental effect, causal effect.}. From the advertiser's perspective, this leads to efficient utilization of marketing budget. From the customer's perspective, this potentially introduces new programs instead of the same recommendations time and time again. There are three key challenges in addressing the causal effects of marketing:
\begin{itemize}
\item Counterfactual estimation - While we know the outcome of a campaign targeted at a particular customer, we also need to know the counterfactual outcome (what would have been) if a different campaign was presented instead. As it is impossible to get the ground truth for counterfactuals, the true incremental effect cannot be measured directly.
\item Unknown total effect size - Marketing campaigns may drive multiple actions or induce repeat behavior.  For example a Fashion ad may lead to a halo effect inducing a customer to purchase multiple accessories in addition to the advertised product.
\item Fast experimentation - Most marketing campaigns are short-lived. We need to quickly estimate the incremental effect and optimize campaign allocation. 
\end{itemize}

We explore a \emph{causal bandit} idea that optimizes campaign allocation based on estimated incremental effects. This approach builds on strengths of causal inference, uplift modeling, and multi-armed bandits. We use the probabilistic exploration of Thompson sampling  for materializing counter-factual outcomes which are used to estimate and optimize context level incremental effect of campaigns. 

%% file: RelatedWork.tex
\section{Related Work}\label{section:related} 

\subsection{Causal Treatment Effect Estimation} 
We first specify the causal treatment effect estimation problem. Let $i = 1, \dots, N$ denote a customer. Suppose we have only one treatment and one control. Let $W_i = \{0, 1\}$ be a binary treatment indicator, with $W_i=1$ denoting that customer $i$ received the treatment and $W_i = 0$ denoting that customer $i$ received the control. Let $Y_i^{(obs)}$ be the observed customer action outcome (e.g.\ click, revenue, other business metric). Let $X_i$ represent the context, potentially including customer features. Each user has a pair of potential outcomes $(Y_i(0), Y_i(1))$. Following the Rubin Causal Model~\cite{Rubin1974,imbens2015}, the customer-level treatment effect $\tau_i$ is defined as the difference in potential outcomes:
\begin{equation}\label{eqn:tau}
\tau_i = Y_i(1) - Y_i(0).
\end{equation}
Dropping suffix $i$ for brevity, let $\Delta$ denote the random variable corresponding to the difference in potential outcomes. Following \citeauthor{athey-imbens} we define the Conditional Average Treatment Effect (CATE) for a given context $\mathbf{x}$ as:we need to
\begin{equation}\label{eqn:taux}
\begin{split}
&\tau(\mathbf{x}) = \mathbb{E}[Y(1) - Y(0) | X = \mathbf{x}] = \\
&\; \int \Delta Pr(Y(1)-Y(0)=\Delta|X=\mathbf{x})  \mathrm{d}\Delta.
\end{split}
\end{equation} 
Say unconfoundedness (selection on observables) assumption holds, which is satisfied in randomized experiments or when the treatment selection is completely determined  through observed $X$. Then, treatment assignment $W_i$ is independent of the potential outcomes given $X$:
\begin{equation}\label{eqn:unconfoundedness}
W \bigCI (Y(0), Y(1)) | X.
\end{equation} 

Prior work in machine learning based estimation of binary CATE can be categorized in three model types \cite{athey-imbens, PredictedLift}\footnote{\citeauthor{athey-imbens} specify decision-tree models. Since the core ideas apply to general supervised learning approaches, we take the liberty of paraphrasing.}:
\begin{itemize}
\item Single-model - As the name suggests, this approach trains one model to predict $Y$ as a function of $W$ and $X$ covariates. Let $\mu(w, \mathbf{x}) = \mathbb{E}[Y^{(obs)}|W = w, X = \mathbf{x}]$, and $\hat{\mu}$ its estimator. Given context vector $\mathbf{x}$, the estimator is scored twice using potential values for treatment $(0, 1)$, and CATE estimated as their difference:
\begin{equation}\label{eqn:singleModel} 
\hat{\tau}(\mathbf{x}) = \hat{\mu}(1, \mathbf{x}) - \hat{\mu}(0, \mathbf{x}).
\end{equation}
\item Two-model - This approach trains two models $\hat{\mu_T}$ and $\hat{\mu_C}$ to separately estimate $Y$ on treatment and control customers. Given context vector $\mathbf{x}$, both estimators are scored once and CATE estimated as the difference:
\begin{equation}\label{eqn:dualModel} 
\hat{\tau}(\mathbf{x}) = \hat{\mu_T}( \mathbf{x}) - \hat{\mu_C}(\mathbf{x}).
\end{equation}
\item Transformed outcome model - This approach uses the insight that if the observed outcome is transformed as:
\begin{equation}\label{eqn:tot}
\begin{split}
\tilde{Y} = & Y(1) \times \frac{W}{Pr(W = 1|\mathbf{x})}\\ 
&\,- Y(0) \times \frac{ 1 - W}{1 - Pr(W = 1|\mathbf{x})}
\end{split}
\end{equation}
then $\mathbb{E}[\tilde{Y} | \mathbf{x}]$ is an unbiased estimator of CATE $\tau(\mathbf{x})$. 
One transforms all outcomes, and uses conventional regressors to directly predict CATE. 
\end{itemize}

An important shortcoming of single and two model approaches is the lack of an appropriate goodness-of-fit or cross-validation measure for model selection,  as there is no ground truth for the conditional average treatment effect $\tau(\mathbf{x})$ \cite{NIPS2007_3248}. The transformed outcome method discards information, and it may be more efficient to estimate $\tau(\mathbf{x})$ using triplet $(Y^{(obs)}, W, X)$ than only $(\tilde{Y}, X)$. 


\citeauthor{athey-imbens} finally present a new algorithm: Causal Tree. They refine the tree construction algorithm to obtain an unbiased estimator of $\tau(\mathbf{x})$. In particular the average treatment effect is estimated by averaging the treatment effect from tree leaves. Units clustered into decision tree leaves are considered synthetic twins of each other, using which CATE within a leaf is estimated. \citeauthor{shalit2017} present theoretical analysis and a family of algorithms based on representation learning that estimates CATE (or individual treatment effects) from observational data. Their algorithm learns a balanced representation such that the induced treatment and control distributions look similar. 

\subsection{Uplift Modeling}
Uplift modeling is a marketing technique to measure the effectiveness of a marketing action and predict its incremental response~\cite{TrueLiftModel, IncrementalValueModeling}. Similar to causal methods, the simple two-model uplift approach models two populations separately and subtracts their lift~\cite{PredictedLift}. For continuous models, two-model uplift is defined as: 
\begin{equation}
\mathbb{E}(outcome|treatment) - \mathbb{E}(outcome|control).
\end{equation} 

The quality of an uplift model is often evaluated by computing an uplift curve~\cite{SignificanceBasedUpliftTrees, DPSVM}, similarly to how an ROC curve is constructed~\cite{nassifROC}. First, we defines a model lift measure over the ranked customers. A lift example is the cumulative outcome amongst the model's top $\rho$ fraction of ranked examples. We then plot the uplift curve by variying $0 \leq \rho \leq 1$ in:
\begin{equation}\label{uplift1}
\hbox{Uplift}(\rho) = \hbox{Lift}(\rho, \hbox{treatment}) - \hbox{Lift}(\rho, \hbox{control}).
\end{equation} 
Uplift curves capture the additional outcome obtained due to the treatment, and is sensitive to variations in coverage~\cite{SAYL}. The higher the uplift curve, the more profitable a marketing model/intervention is.  

\subsection{Multi-Armed Bandits} 
We need to predict the incremental effect of future campaigns when presented to potentially new customers. Marketing campaigns are transient, share common features, and need to be tested quickly~\cite{Hill:2017:EBA:3097983.3098184}. Multi-armed bandit algorithms are effective for rapid experimentation because they concentrate testing on treatments that have the greatest potential reward~\cite{bubeck2012BanditsSurvey, Hill:2017:EBA:3097983.3098184}. They optimally balance exploration and exploitation to achieve minimal regret, and incorporate customer context to personalize prediction~\cite{Agrawal2013ContextualTSBounds, Dani2008ContextualBanditBound}.

Thompson sampling is a common bandit algorithm where a campaign is selected proportionally to the probability of that campaign being optimal, conditioned on previous observations. In practice, this probability is not sampled directly. Instead one samples model parameters from their posterior and picks the campaign that maximizes the reward~\cite{chapelle2011empirical}.

The standard bandit setting assumes unconfoundedness, although non-contextual bandits with unobserved confounders have been developed~\cite{BanditsUnobservedConfounders}. In the case of known confounders, one may resort to inverse propensity weighing and other off-policy policy evaluation methods to unbias the estimates~\cite{Li:2015:IPS, Swaminathan:2015:POEM}.

\subsection{Multi-Campaign Setting} 
In this study we focus on general multi-campaign settings where $k>2$ marketing campaigns can be active at any time. Let $W = \{0, 1, \ldots, K\}$ be the campaign indicator with $W = k$ if customer received campaign $k$. Let $W = 0$ be a special case (customer held out of marketing or receives control). Let $H$ denote the binary conversion indicator for some action that we want to drive through these marketing campaigns, and $Y$ its business metric. Then the estimated CATE of campaign $k$ becomes (see proof in Appendix~\ref{section:appendix}):
\begin{equation}\label{eqn:incrRevenueType1}
\begin{split}
\hat{\tau}_k(\mathbf{x}) =  & (Pr( H = 1 | X = \mathbf{x}, W = k)\\
&\;- Pr( H = 1 | X = \mathbf{x}, W = 0)) \\
& \times  (  \mathbb{E}[Y | X = \mathbf{x}, H = 1] -  \mathbb{E}[Y | X = \mathbf{x}, H = 0] ).
\end{split}
\end{equation}

One can estimate CATE's first term (incremental propensity) in three steps: a) build a conversion model on customers exposed to a given campaign; (b) build a conversion model on customers not exposed to that campaign; (c) score both models on a new customer and take their score difference. One can also use predictive models to approximate Eqn.~\ref{eqn:incrRevenueType1}'s second term, business metric difference. This two-model approach has limitations:
\begin{itemize}
\item Requiring a M:1 mapping from campaigns to actions is limiting. A campaign may drive more than one action and its total incremental effect should include all its actions. Attribution can be challenging if multiple campaigns are involved. If we allow a M:N mapping between campaigns and actions, a campaign's incremental effect becomes more comprehensive but computationally cumbersome:
\begin{equation}\label{eqn:incrRevenueType2}
\begin{split}
&\hat{\tau}_k(\mathbf{x}) =\\
&\sum_j  (  Pr( H^j = 1 | X = \mathbf{x}, W = k) \\
&\quad\quad - Pr( H^j = 1 | X = \mathbf{x}, W = 0))\\
&\times ( \mathbb{E}[ Y | X = \mathbf{x}, H^j = 1] - \mathbb{E}[ Y | X = \mathbf{x}, H^j = 0] ).
\end{split}
\end{equation}
\item 
Customer response to marketing is impacted by content quality and environmental factors (time of day, day of week, device). It would be ideal to compute campaign or context-specific incremental propensity.
\item Getting a pure control group ($W_i = 0$) is challenging. A control customer may receive substitute campaigns via other marketing channels. It is difficult to account for such exogenous factors. 
\item  Uplift modeling approaches that model uplift directly tend to outperform their two-model counterparts~\cite{SignificanceBasedUpliftTrees, RDP}.
\end{itemize}


%% file: CausalProposal.tex
\section{Causal Effect Prediction Proposal}\label{section:incremental}
Our approach draws on the strengths of causal inference, uplift modeling, and multi-armed bandits. Akin to causal trees, we optimize on causal treatment effect rather than pure outcome, and incorporate counterfactual matching within data collection. Following uplift modeling results, we directly optimize the incremental business metric. We use contextual bayesian multi-armed bandits to scale to multiple treatments and to perform off-policy evaluation on logged data. In particular, Thompson sampling enables exploration of treatments on similar customers and materialization of counterfactual outcomes. 

\subsection{Problem Formalization}
As we are dealing with multiple treatments, we generalize the customer-level treatment effect to:
\begin{equation}\label{eqn:tauMT}
\tau_k = Y(W=k) - Y(W \ne k),
\end{equation}
where $(Y(W=k), Y(W \ne k))$ are the postulated outcomes of receiving treatment $k$ and not receiving treatment $k$, respectively. We define: 
\begin{equation}
\begin{split}
&Y(W \ne k) = \\
&\int\displaylimits_y y \sum_{m \ne k} Pr(Y = y | X, W=m) Pr(W=m) \mathrm{d}  y.
\end{split}
\end{equation}

Similarly, we define the treatment-specific Conditional Average Treatment Effect (CATE) for campaign $k$ as:
\begin{equation}\label{eqn:tauxMT}
\tau_k(\mathbf{x}) = \mathbb{E}[Y(W=k) - Y(W \ne k) | X = \mathbf{x}].
\end{equation} 
The unconfoundedness assumption becomes:
\begin{equation}\label{eqn:unconfoundednessMT}
W \bigCI (Y(0), \dots, Y(K)) | X.
\end{equation} 
Let us define the customer-level treatment-specific incremental effect for customer $i$ as:
\begin{equation}\label{eqn:customerIncrementalEffect}
Y_{i,k}^* = Y_i(W_i=k) - Y_i(W_i \ne k).
\end{equation}
We now present a treatment-specific CATE based on expectation of the difference, rather than the difference in expectations of outcomes from two separate models:
\begin{equation}\label{eqn:finalIncr}
\hat{\tau}_k(\mathbf{x}) = \mathbb{E}[Y_{i,k}^* | X_i = \mathbf{x}].
\end{equation}

The choice of $W \ne k$ baseline in Eqn.~\ref{eqn:tauMT}, which includes the possibility of not showing any treatment, is in the same spirit as one-versus-all multi-class classification. Such an approach is appropriate from a business-perspective, and our experiments show benefit over using non-causal approaches. However, since each arm has a different baseline, one may argue that using a common baseline across all arms may result in a better CATE estimate. For example, one can drop customers who didn't see any treatment, and compare against a constant control arm, or the average performance of all arms. We note that the latter methods conserve the same arm ranking as a non-causal bandit, and argue that preserving non-treatment customers is closer to the production logic. We are exploring these different types of baseline and their trade-offs as part of our future work.

\subsection{Training Data Generation with Incremental Target Calculation}
To measure and predict the incremental effect of treatments $Y^*_{i, k}$, we leverage off-policy evaluation techniques~\cite{Swaminathan:2015:POEM} and require that historical treatments are logged properly. Let logs be $L = \{X_i, W_i, P_i, Y^{(obs)}_i)\}$ where user $i$ with context $X_i$ was shown treatment $W_i = k$ with probability $P_i > 0$ and outcome $Y^{(obs)}_i$. The nature of observed outcome is application-specific, such as clicks, repeat visits, difference in customer spending before and after the experiment, etc. We estimate the counterfactual $Y_i(W_i \ne k)$ based on historical records whose context matches $X_i$.

More formally, let $\Phi(X)$ be a context matching algorithm that finds similar customers to $X$, like K-Nearest Neighbor, locality sensitive hashing, or propensity matching. Given customer $X_i$ with observed outcome $Y^{(obs)}_i$, we use $\Phi(X_i)$ to identify similar customers who were not shown treatment $k$, and use their logged outcomes to estimate a counterfactual outcome $Y_i(W_i \ne k)$ for customer $X_i$. Algorithm~\ref{algo:trainingData} shows our training data generation approach where we leverage similar historical logs to build the incremental effect training data. 

\begin{algorithm}[ht!]
	\caption{Training Data Generation for Incremental Outcome Prediction Model}
	\label{algo:trainingData}
	\begin{algorithmic}
		\STATE {\bfseries Inputs:} Size of training data $M > 0$; logged events $L$; number of matching observations $M'$; context matching algorithm $\Phi$.
		\STATE $T_0 \leftarrow \emptyset$ \COMMENT{An initially empty training data}
		\FOR{$m=1$ {\bfseries to} $M$}
		\STATE Sample record  $(\mathbf{x}_m, w_m, p_m, y_m^{(obs)})$ from $L$.
		\STATE $y_m^{(tr)} \gets y_m^{(obs)}/ p_m$ \COMMENT{Apply policy bias correction to outcome}
		\STATE $y_m^{(cf)} \gets 0$ \COMMENT{Initialize counterfactual outcome to zero}
		\STATE 
Select $M'$ similar records using $\Phi(\mathbf{x}_m, \mathbf{x}_{m'})$
		\FOR{$m'=1$ {\bfseries to} $M'$}  
		        \STATE Pick record $(\mathbf{x}_{m'}, w_{m'}, p_{m'}, y_{m'}^{(obs)})$ with $w_m \ne w_{m'}$ from $L$
			\STATE $y_m^{(cf)} \gets y_m^{(cf)} + y_{m'}^{(obs)} / p_{m'}$	
                 \ENDFOR
                 \STATE $y_m^{(cf)} \gets y_m^{(cf)} / M'$ \COMMENT{Final counterfactual estimate}
                 \STATE $y_{m, w_m}^* \gets y_m^{(tr)} - y_m^{(cf)}$  \COMMENT{Unbiased incremental target} 
		\STATE $T_m \leftarrow $ CONCATENATE($T_{m-1}, (\mathbf{x}_m, w_m, y_{m, w_m}^*)$)
		\ENDFOR
	\end{algorithmic}
\end{algorithm}

Suppose we have historical data logs such that $P(k,X_i)>0, \, \forall k, X_i$. There are no hidden confounders because $W_i$ is chosen with probability $P_i$ and that probability is computed depending only on $X_i$. The outcome $Y_i$ and covariates outside $X_i$ are unknown to the software that evaluates $P_i$, so $Y_i$ must be independent of these. We then use inverse propensity estimation to unbias $Y^*$~\cite{Li:2015:IPS}.  One may further refine the inverse propensity estimator to control for variance \cite{JMLR:v16:swaminathan15a}.

\subsection{Using Thompson Sampling for Counterfactual Materialization}

To ensure online-optimization and a balanced explore-exploit mix, we use multi-armed bandits trained on Algorithm~\ref{algo:trainingData} data to estimate Eqn.~\ref{eqn:finalIncr}. Let $\theta = (\theta_1, \theta_2, \ldots, \theta_K)$ denote the model parameters for the K-armed bandit model. Let $W^{*}(\theta_t,\mathbf{x})$ denote the optimal treatment given context $\mathbf{x}$ and model parameters $\theta$ at time $t$. At any given time $t+1$, Thompson sampling selects a treatment $W=k$ proportionally to the probability of that treatment being optimal:
\begin{equation}
W_{t+1} \sim Pr(W=W^{*}(\theta_t,\mathbf{x})|X = \mathbf{x}).
\end{equation}
When there is no strong evidence that any one treatment is optimal, the bandit explores multiple treatments on similar contexts. As the bandit improves its treatment success estimates, it is more likely to exploit the winning one~\cite{chapelle2011empirical}. As such, Thompson sampling bandits are a perfect fit for the materialization of counterfactual outcomes. The observed contexts and outcomes determine $\theta_t$ and $Pr(W=W^{*})$. We log $P_i = Pr(W_i = W^{*})$, making the proposed framework a closed-loop system~\cite{DBLP:journals/corr/AgarwalBCHLLLMO16}. 

Algorithm~\ref{algo:bandit} specifies the overall contextual bayesian bandit algorithm. We suppose that at time $t$, multiple contexts are received. The increment in $t$ can be understood to correspond to one day where predictions are made on multiple contexts, but model update happens asynchronously after delayed incremental outcomes are computed. 

\begin{algorithm}[ht!]
	\caption{Thompson Sampling based Contextual Multi-Armed Bandits with Online Scoring and Batch Training}
	\label{algo:bandit}
	\begin{algorithmic}
         	\STATE {\bfseries Initialization:} Time $t=0$; event log~\mbox{$L = \{\}$};  d-dimensional bandit arm contextual distribution parameters \mbox{$\theta^k \sim \mathcal{N}_d(0, 1), \forall k$} arms. 
		\FOR{$t=1, 2, \ldots$} 
		\STATE \COMMENT{Online Scoring}
		\FOR{$m= 1, 2, \ldots$}
		\STATE Receive d-dimensional context $\mathbf{x}_m$. 
		\FOR{Arm $k = 1, 2, \ldots, K$}
		\STATE Sample $\theta \sim \theta_k$ 
		\STATE Estimate $y^*_{m,k} = \theta^T\mathbf{x}_m.$
		\ENDFOR 
		\STATE Play arm $w_m = \argmax_k y^*_{m,k}$.
		\STATE Log $(x_m, w_m, p_m)$. 
		\ENDFOR
		\STATE \COMMENT{Batch Training}
		\STATE Asynchronously update log $L$ with delayed reception of new rewards $Y_m^{(obs)}$ for $m = 1, 2, ...$.
		\STATE Update $\theta_k, \forall k$ using training data generated by Algorithm~\ref{algo:trainingData}.
		\ENDFOR 
	\end{algorithmic}
\end{algorithm}

%% file: Experiments.tex
\section{Experiments}
\begin{figure}
\centering
\subfigure{
\includegraphics[width=.225\textwidth]{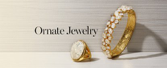}
}
\subfigure{
\includegraphics[width=.225\textwidth]{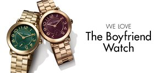}
}
\subfigure{
\includegraphics[width=.225\textwidth]{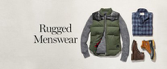}
}
\subfigure{
\includegraphics[width=.225\textwidth]{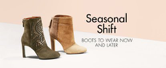}
}
\caption{Examples of Fashion campaigns used in data collection.}
\label{fig:creatives}
\end{figure}

Experiments in this paper are based on offline evaluation on marketing logs. 
We collected an Amazon Fashion marketing dataset where 410K randomly sampled treatment customers were randomly targeted with one of 16 marketing campaigns upon visiting a retail app. These campaigns spanned across men's and women's fashion in clothing, shoes, jewelry, and watches (examples shown in Figure~\ref{fig:creatives}). Separately, we randomly sampled a control hold-out set of 100K customers that did not receive any campaign from our experiment (customers may have seen other business-as-usual messages). The resulting dataset was split into 70:30 training and testing. Following the concepts of off-policy policy evaluation, we only consider targeted customers who  were shown the same campaign as the new model prediction~\cite{Li:2015:IPS}. 

As incrementality measures are often continuous, our Algorithm~\ref{algo:bandit} needs to be a contextual bayesian bandit that optimizes a continuous target metric. We use a Bayesian Linear Probit regression extension of~\cite{graepel2010BLIP} that has a continuous support. We use this regression as the basis of a Bayesian Generalized Linear Bandit, as done in~\cite{Teo2016airstream} and~\cite{Hill:2017:EBA:3097983.3098184}. 

We featurized campaign content and customer behaviors as well as their interactions. We winsorized continuous features to reduce the effect of outliers, applied log-transformation to spend and pricing related features, and an overall min-max scaling to limit features to a [0,1] range. Given the large number of features and campaigns compared to the size of available training data, we performed feature selection prior to building predictive models. We chose the $S$ best features ($S \in \{100, 200, 300, \ldots\}$) based on the regression feature-target F-value between features and the continuous target variable. We then trained a bandit model on each feature set $S$, and measured its cross-validated performance to identify the best model to test.

Incremental outcomes ($Y^*$) are generated using Algorithm~\ref{algo:trainingData}, as the difference in the business metric between the targeted customer and $M'$ similar customers who were not shown that specific campaign (or any campaign). As a context matching algorithm $\Phi$, we used a GPU-based nearest-neighbor search service that uses the \emph{hierarchically navigable small worlds} similarity algorithm~\cite{hnsw} to match normalized customer feature vectors. We tried two different values of $M' \in \{10, 50\}$ both of which had similar results (Pearson correlation 0.92, p-value 0.0). We fix $M'=10$ in further experiments.

%% file: Results.tex
\section{Results}
\subsection{Causal Bandit Performance}
We first compare two types of models: 
\begin{description}
\item[Non-incremental model] uses the bayesian probit regression bandit to optimize the total value of a fashion business metric of the targeted customers.
\item[Incremental model] is the the proposed causal bandit model. It uses the bayesian probit regression bandit to optimize the incremental value of the same fashion business metric of the targeted customers, wherein increment is computed using Eqn.\ref{eqn:customerIncrementalEffect} and Algorithm~\ref{algo:trainingData}.
\end{description} 

For each customer, each bandit model scores all campaigns, and recommends the campaign with the highest score. We train each model on the treatment training set, and use the trained model to score and recommend campaigns for both the treatment testing set, as well as the control hold-out set.

To visualize the treatment effect, we rank the testing and hold-out customers by their recommended campaign predictive score, and divide the ranked list into ten deciles. Decile 10 represents the top $10\%$ campaign-susceptible customers and decile 1 represents the bottom $10\%$. Each decile contains a segment of targeted customers shown the same campaign as recommended by the scoring model and a segment of hold-out customers. We refer to these segments as the decile-level treatment and control groups. The decile treatment effect is the differrence between the observed business metric for treament and control decile-level segments (recall customers were randomly assigned to treatment and control).

\begin{figure}
	\centering
	\includegraphics[width=\columnwidth]{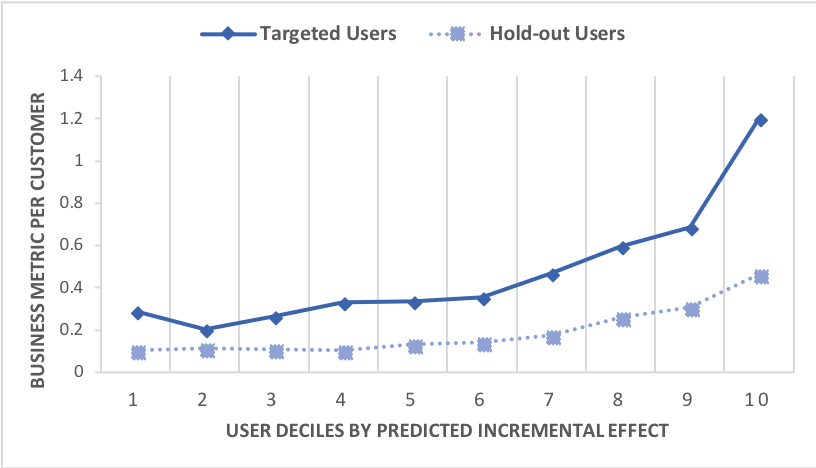}
	\caption{Per-customer business metric per decile, measured on the treatment and control sets. Deciles are ranked using the incremental model. The difference between curves is the causal effect of marketing in that decile. }
	\label{fig:averageBusinessMetricByDecile}
\end{figure}

Figure~\ref{fig:averageBusinessMetricByDecile} uses the decile ranking produced by the incremental model, and plots the actual business metric average per decile \footnote{The business metric is observed in a future time period post targeting.}. For the top decile, incremental-model targeting results in $1.2$ business metric units, compared to $0.46$ units for control. The difference of $0.74$ between these two values is precisely the conditional average treatment effect (CATE) in that decile. Analyzing all deciles, we observe a good correlation between predictive incremental bandit score and the actual causal effect, thus highlighting the model's utility in making useful advance predictions for business decisions. 

Readers may wonder about the oddity that decile 2 has a smaller score than decile 1. We believe this is caused by new-to-site customers who are placed in decile 1 due to lack of historical behaviors (e.g. model confuses them for inactive customers and assigns a low score). However, these customers are actually responsive and raise the per-customer performance in decile 1 compared to decile 2. 

\begin{figure}[th]
	\centering
	\includegraphics[width=\columnwidth]{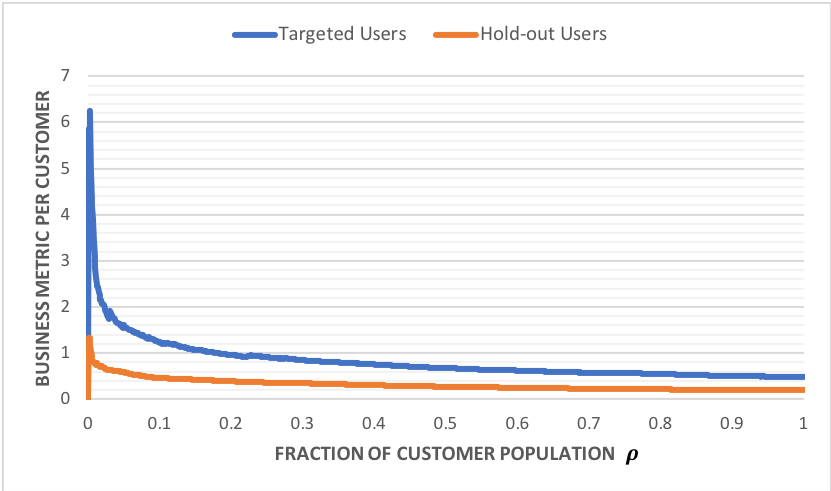}
	\caption{Causal model lift curves for targeted and hold-out customers. The difference between the curves at point $\rho$ on X-axis is the uplift in business metric at that point.}
	\label{fig:liftCurves}
\end{figure}

Budget limitations typically force business to limit the number of customers exposed to different marketing activities. This suggests that business teams will place more importance in top-ranked customers and desire a segment of top-ranked customer which yields better cumulative return on investment. This can be represented as an uplift curve (Eqn.~\ref{uplift1}). A point $\rho \in [0, 1]$ on the X-axis of Figure~\ref{fig:liftCurves} corresponds to the top $\rho$ fraction of customers as sorted by decreasing order of bandit score. We normalize lift to represent the per-customer average outcome amongst the model's top $\rho$ fraction of ranked examples.

\begin{figure}
\centering
\subfigure[Entire population]{
\includegraphics[width=\columnwidth]{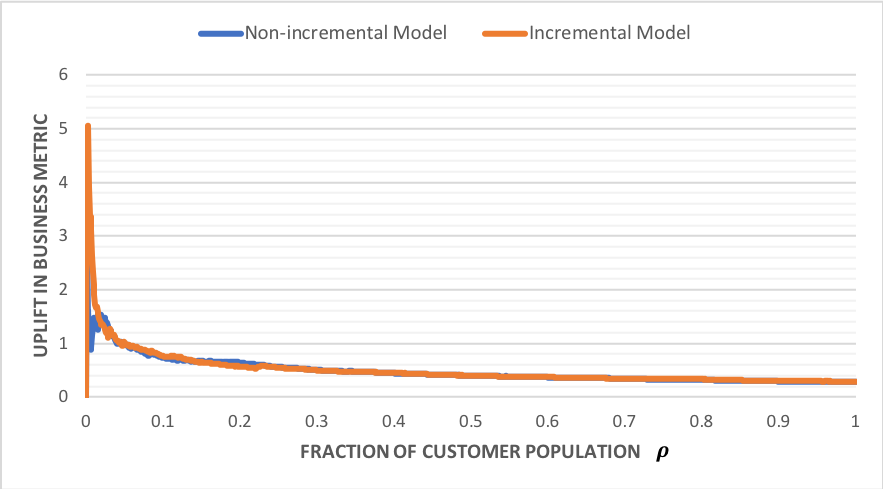}
\label{fig:upliftIcml}
}\\ 
\subfigure[Top 10\% ranked population]{
\includegraphics[width=\columnwidth]{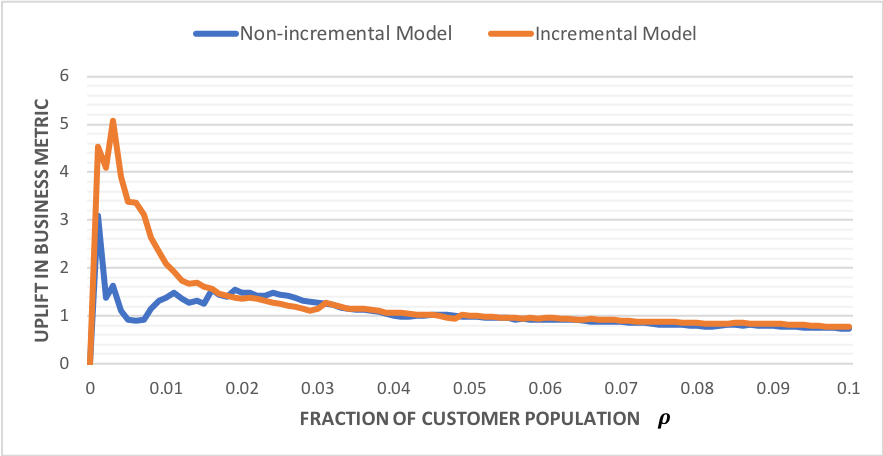}
\label{fig:uplift10pctIcml}
}
\caption{Comparing uplift in business metric by incremental and non-incremental approaches.}
\label{fig:upliftCurves}
\end{figure}

The difference between the two normalized lift curves for treatment (targeted) and control (hold-out) customers is the uplift curve, plotted in Figure~\ref{fig:upliftIcml}. The incremental model dominates almost entirely, and uplift differences can be better distinguished in the top 10\% targetable population (Figure~\ref{fig:uplift10pctIcml}), stressing the importance of modeling incrementality. The dip at zero is natural because there is zero total reward to be obtained without selecting any customers. 

To understand the relationship between Figures~\ref{fig:averageBusinessMetricByDecile} and~\ref{fig:liftCurves}, one can compare the performance on targeted customers in both figures. We see that the point estimate at $\rho = 0.1$ of Figure \ref{fig:liftCurves} is the same as the per customer business metric in decile 10 of Figure \ref{fig:averageBusinessMetricByDecile}, as both measure the average value of the business metric over the targeted 10\% customers with highest predictive scores. The point estimate at $\rho = 0.2$ in Figure~\ref{fig:liftCurves} is the average of estimates in deciles 9 and 10 of Figure~\ref{fig:averageBusinessMetricByDecile} (top 20\% customers) and so on. In other words, Figures~\ref{fig:averageBusinessMetricByDecile} and~\ref{fig:liftCurves} represent the non-cumulative and cumulative versions of the same business metric, corresponding to CATE and uplift respectively. 

At $\rho=0.01$, the incremental model has an uplift of $209$ units of business metric, compared to $108$ units for the non-incremental model. At $\rho=0.1$, the incremental model has an uplift of $74$ units of business metric compared to $70$ units for the non-incremental model. For the whole population $\rho=1$, the uplift values are $28.2$ and $26.7$ units respectively. Such high values, specifically for the top percentile, showcase the potential of our method. 

\subsection{KNN Context Matching}
To test the effect of our counterfactual estimation, we replace the KNN-based CATE estimator of incremental model by  a two-model alternative extended to multiple campaigns. For the latter, we train two separate linear regression models to predict the business metric for treatment and control customers ($\hat{\mu_T}$ and $\hat{\mu_C}$ in Eqn.~\ref{eqn:dualModel}). After scoring each customer with both models, we determine the campaign with the highest difference $\hat{\mu_T}(\mathbf{x}) - \hat{\mu_C}(\mathbf{x})$. Using the same offline evaluation technique as before, we determine and plot the uplift curve. 

\begin{figure}
	\centering
	\includegraphics[width=\columnwidth]{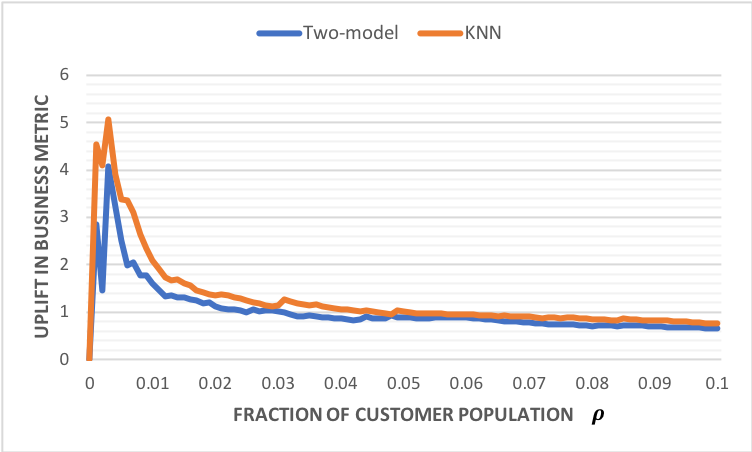}
	\caption{KNN-based approach to CATE estimation yields better uplift than the two-model approach overall, particularly in the 10\% of scored population.}
	\label{fig:mtmc10}
\end{figure}

The KNN uplift curve tends to dominate its two-model counterpart especially in the top 10\% (Fig.~\ref{fig:mtmc10}), where the two-model performance fluctuates. We hypothesize two reasons as to why KNN leads to superior incremental effects. First, the KNN approach determines causal effect with respect to other marketing campaigns, while the two-model approach compares only to a no-targeting control option. Thus, KNN approach may be more comprehensive in counterfactual estimation. Second, KNN can represent complex non-linear relationships while the current two-model approach is limited to linear. 

%% file: Conclusion.tex
\section{Conclusion and Future Work}
We present a multi-armed bandit approach that optimizes advertising campaign targeting based on incremental or causal outcomes. We present proof-of-concept results using an offline fashion marketing dataset. We compare our causal approach to non-causal alternatives, and observe that our approach dominates in terms of incremental outcomes in targeted customers over a random hold-out group.

We are currently investigating a few improvements. First, we plan to utilize a consistent baseline across all marketing campaigns instead of varying baselines $(W_i \ne k)$ in Section \ref{section:incremental}. We are also interested in explicitly modeling the trade-off between short-term and long-term objectives as a composite objective based on a weighted average, like:
$\beta \times P(click) \times \mathbb{E}(metric|click) + (1-\beta) \times \mathbb{E}(incremental\_metric|impression)$. This is important in developing generic campaign management frameworks where the notion and importance of short-term and long-term objectives may be different.  Finally, we plan on deploying our system at a larger scale to enable further experimentation, assess impact and fine tune our methodology. 


%% file: Appendix.tex
\section{Appendix}\label{section:appendix}
\begin{proof}
We now prove Eqn.~\ref{eqn:incrRevenueType1}. Let $H$ denote the binary conversion indicator for the action we want to drive, then:
\begin{equation}
\label{eqn1Repeat}
\begin{split}
\mathbb{E}[Y | X& = \mathbf{x}, W = k]\\
= & \int \! y \times Pr(y |X = \mathbf{x}, W = k)\, \mathrm{d}y \\
 = & \int \! y \times [\sum_{h \in \{0, 1\}} Pr(y, H= h |X = \mathbf{x}, W = k)] \mathrm{d}y  \\
 = & \int \! y \times  [ \sum_{h \in \{0, 1\}}  Pr(y|H=h,X=\mathbf{x},W=k) \\ 
& \times Pr(H=h|X=\mathbf{x},W=k) ] \mathrm{d}y \\
= & \sum_{h \in \{0, 1\}} [ Pr(H=h|X=\mathbf{x},W=k) \\
& \times \int \! y \times Pr(y|H=h,X=\mathbf{x},W=k)\, \mathrm{d}y ] \\
= & \sum_{h \in \{0, 1\}} [ Pr(H=h|X=\mathbf{x},W=k) \\
& \times \int \! y \times Pr(y|H=h,X=\mathbf{x})\, \mathrm{d}y ].
\end{split}
\end{equation}

The last step is simplified based on the assumption that $Y$ is conditionally independent of $W$ given $H$ and $X$. For $W=0$, we can similarly show that:
\begin{equation}
\label{eqn2Repeat}
\begin{split}
\mathbb{E}[Y  | X& = \mathbf{x}, W = 0]  = \\
& \sum_{h \in \{0, 1\}} [ Pr(H=h|X=\mathbf{x},W=0) \\
& \times \int \! y \times Pr(y|H=h,X=\mathbf{x})\, \mathrm{d}y ].
\end{split}
\end{equation}

Recall binary CATE: $\hat{\tau}_k(\mathbf{x}) = E(Y  | X = \mathbf{x}, W = k) - E(Y  | X = \mathbf{x}, W = 0)$. Subtract Eqn.~\ref{eqn2Repeat} from Eqn.~\ref{eqn1Repeat}:
\begin{equation}
\begin{split}
&\hat{\tau}_k(\mathbf{x}) =  \int \! y \times Pr(y|H=1,X=\mathbf{x})\, \mathrm{d}y \\
& \times [Pr(H=1|X=\mathbf{x},W=k) - Pr(H=1|X=\mathbf{x},W=0)] \\
& + \int \! y\times Pr(y|H=0,X=\mathbf{x})\, \mathrm{d}y \\
& \times [Pr(H=0|X=\mathbf{x},W=k) - Pr(H=0|X=\mathbf{x},W=0)]
\\
\end{split}
\end{equation}
Simplifying and re-arranging, 
\begin{equation}
\begin{split}
&\hat{\tau}_k(\mathbf{x}) =  \\
& \left(Pr( H = 1 | X = \mathbf{x}, W = k) - Pr( H = 1 | X = \mathbf{x}, W = 0) \right) \\
& \times  (  \mathbb{E}[Y | X = \mathbf{x}, H = 1] -  \mathbb{E}[Y | X = \mathbf{x}, H = 0] ).
\end{split}
\end{equation}
\end{proof}